\documentclass[a4paper, 12pt]{article}
\pdfoutput=1
\usepackage[utf8]{inputenc}
\usepackage[titlepage,sectionmark]{polytechnique}
\usepackage[T1]{fontenc}
\usepackage{amsmath,amsfonts,amssymb}
\usepackage{bm}
\usepackage{algorithm}
\usepackage{algorithmic}
\usepackage{graphics}
\usepackage{hyperref}
\usepackage{dirtytalk}
\usepackage{listings}

\title{Internship report}
\subtitle{Meta-learning algorithms for Few-Shot Computer Vision}
\author{Etienne \textsc{Bennequin} \\ 
\vspace{5mm}
\large{Under the supervision of Antoine Toubhans, PhD}}
\date{April - July 2019}
\logo{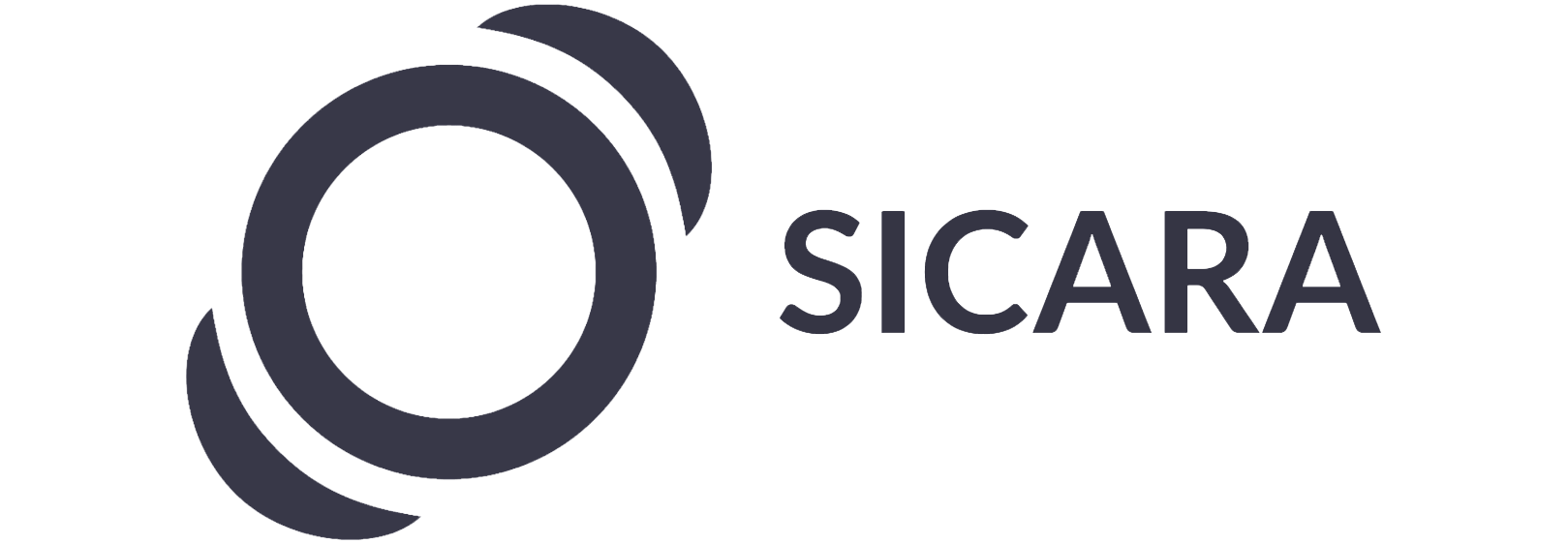}
\begin{document}

\maketitle

\section*{Abstract}
Few-Shot Learning is the challenge of training a model with only a small amount of data. Many solutions to this problem use meta-learning algorithms, \emph{i.e.} algorithms that learn to learn. By sampling few-shot tasks from a larger dataset, we can teach these algorithms to solve new, unseen tasks.

This document reports my work on meta-learning algorithms for Few-Shot Computer Vision. This work was done during my internship at Sicara, a French company building image recognition solutions for businesses. It contains:
\begin{enumerate}
    \item an extensive review of the state-of-the-art in few-shot computer vision;
    \item a benchmark of meta-learning algorithms for few-shot image classification;
    \item the introduction to a novel meta-learning algorithm for few-shot object detection, which is still in development.
\end{enumerate}

\vspace{80pt}
\section*{Special thanks}
I would like to thank everyone at Sicara for their help on so many levels. In particular, thanks to Antoine Toubhans, Elisa Negra, Laurent Montier, Tanguy Marchand and Theodore Aouad for their daily support.

\newpage
\tableofcontents

\newpage
\section{Introduction}
In 1980, Kunihiko Fukushima developed the first convolutional neural networks. Since then, thanks to increasing computing capabilities and huge efforts from the machine learning community, deep learning algorithms have never ceased to improve their performances on tasks related to computer vision. In 2015, Kaiming He and his team at Microsoft reported that their model performed better than humans at classifying images from ImageNet~\cite{he2016deep}. At that point, one could argue that computers became better than people at harnessing billions of images to solve a specific task. 

However, in real world applications, it is not always possible to build a dataset with that many images. Sometimes we need to classify images with only one or two examples per class. For this kind of tasks, machine learning algorithms are still far from human performance.

This problem of learning from few examples is called few-shot learning.

For a few years now, the few-shot learning problem has drawn a lot of attention in the research community, and a lot of elegant solutions have been developed. An increasing part of them use meta-learning, which can be defined in this case as \emph{learning to learn}.

During my internship at Sicara, I focused on meta-learning algorithms to solve few-shot computer vision tasks, both for image classification and object detection. I compared the performance of four distinct meta-learning algorithms in few-shot classification tasks. I also started the development of a novel meta-learning model for few-shot object detection.

The first section is an extensive review of state-of-the art solutions for solving few-shot image classification and few-shot image detection. It starts with the definition of the few-shot learning problem.

Then I will expose my contributions. The first part of it is a benchmark of state-of-the-art algorithms for few-shot image classification on several settings and datasets. The second part introduces the YOLOMAML, a novel solution for few-shot object detection. This algorithm is still in development.

This report shares details about the research process and the implementation of the algorithms and experiments. I hope this information about the issues raised during my work and my attempts at solving them will be useful for anyone who will work on meta-learning algorithms in the future.

\newpage
\section{Review}
\subsection{Few-Shot classification problem} \label{few-shot-classification}
\begin{figure}[h]
    \centering
    \includegraphics[width=10cm]{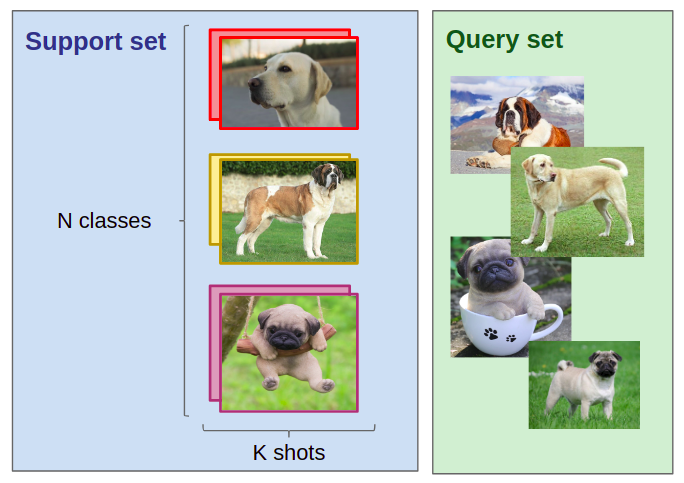}
    \caption{A $3$-way $2$-shot classification problem. Images from the query set would need to be classified in $\lbrace$ Labrador, Saint-Bernard, Pug $\rbrace$.}
    \label{fig:FSCP}
\end{figure}

We define the $N$-way $K$-shot image classification problem as follows. Given:
\begin{enumerate}
    \item a support set composed of:
    \begin{itemize}
        \item $N$ class labels;
        \item For each class label, $K$ labeled images;
    \end{itemize}
    \item $Q$ query images;
\end{enumerate}
we want to classify the query images among the $N$ classes. The $N \times K$ images in the support set are the only examples available for these classes.

When $K$ is small (typically $K<10$), we talk about \emph{few-shot image classification} (or \emph{one-shot} in the case where $K=1$). The problem in this case is that we fail to provide enough images of each class to solve the classification problem with a standard deep neural network, which usually require thousands of images. Note that this problem is different from semi or weekly supervised learning, since the data is fully labeled. The problem here is not the scarcity of labels, but the scarcity of training data. 

A visual example of a few-shot classification problem is shown in Figure \ref{fig:FSCP}.

The Few-Shot Learning problem (which includes few-shot image classification) has drawn a lot of attention in the past few years. Many different ways of solving this problem have been imagined. They all have in common that they use additional information from a large \emph{base-dataset}. The classes in the base-dataset are different from the ones in the support set of the few-shot task we ultimately want to solve. For instance, if the target task is classifying images as Labrador, Saint-Bernard or Pug (Figure \ref{fig:FSCP}), the base-dataset can be composed of many other dog breeds. Here I provide an overview of these solutions. 

\paragraph{Memory-augmented networks} Santoro \emph{et al.} (2016)~\cite{Santoro16} had the idea that new images from previously unseen classes could be classified by using stored information about previous image classification. Their model uses a Recurrent Neural Networks that learns both how to store and how to retrieve relevant information from past data. Other methods exploit the idea of extending neural networks with external memory~\cite{Munkhdalai17}~\cite{Sprechmann18}.

\paragraph{Metric learning} Koch \emph{et al.} (2015)~\cite{Koch15} proposed the Siamese Neural Networks to solve few-shot image classification. Their model is composed of two convolutional neural networks with shared weights (the \emph{legs}), that compute embeddings (\emph{i.e.} features vectors) for their input images, and one \emph{head} that compares the respective output of each leg. At training time (on the large \emph{base-dataset}), the network receives couples of images as input, predicts whether they belong or not to the same class, and is trained upon the accuracy of this prediction. Ultimately, when evaluated on a few-shot classification class (see Figure \ref{fig:FSCP}), each query image is compared to every images in the support set, and is assigned to the class that is considered the closest (using for instance $k$-Nearest Neighbours). \\
This algorithm achieved interesting results on few-shot image classification. However, the task upon which it was trained (comparison of two images) differed from the task upon which it was evaluated (classification). 

Vinyals \emph{et al.} (2016)~\cite{Vinyals16} considered that this was a drawback and proposed a slightly different version of this algorithm, inside of the \emph{meta-learning framework} (see the definition of this framework in section \ref{meta-learning}). Their Matching Networks also classify query images by comparing their embedding to the embeddings computed from support set images, but the difference is that their training objective is image classification as well. They outperform Siamese Networks, thus validating their assumption.

Later works aim at improving this algorithm~\cite{Snell17}~\cite{Sung18}. They will be presented with more details in section \ref{metric-learning}.

\paragraph{Gradient-based meta-learners} Other algorithms inside of the meta-learning framework learn an efficient way to fine-tune a convolutional neural network on the support set in order to accurately classify the query set. Finn \emph{et al.} (2017)~\cite{Finn17} developed a Model-Agnostic Meta-Learner (MAML) which tries to learn the best parameters for the CNN's initialization in order to achieve good accuracy on the query set after only a few gradient descents on the support set. The Meta-SGD developed by Li \emph{et al.} (2017)~\cite{li2017meta} goes further: in addition to the initialization parameters, this algorithm learns for each parameter a learning rate and an update direction. Ravi \& Larochelle (2016)~\cite{Ravi16} proposed a Long-Short-Term-Memory network where the cell state (\emph{i.e.} the variable supposed to carry long-term memory in a LSTM) is the parameters of the CNN. This allows to execute a \emph{learned gradient descent}, where all the hyper-parameters of the CNN's training are actually trained parameters of the LSTM.

Still inside the meta-learning framework, which they considered as a sequence-to-sequence problem, Mishra \emph{et al.} (2018~\cite{Mishra18} combine temporal convolutions with causal attention to create their Simple Neural AttentIve Learner (SNAIL). Finally, Garcia \& Bruna~\cite{Garcia18} proposed to use graph neural networks as an extension of all meta-learning algorithms for few-shot learning.

\paragraph{Data generation} An other option to solve the problem of having too few examples to learn from is to generate additional data. Hariharan \& Girshick (2017)~\cite{hariharan2017low} augmented metric learning algorithm with hallucinated feature vectors which were added to the feature vectors extracted from real images. Antoniou \emph{et al.} (2017)~\cite{antoniou2017data} applied Generative Adversarial Networks to Few-Shot data augmentation: their GAN are able to take as input an image from a previously unseen class to generate new images which belong in the same class. Wang \emph{et al.} (2018)~\cite{wang2018low} proposed a meta-learned imaginary data generator which can be trained in an end-to-end fashion with a meta-learning classification algorithm.

\paragraph{}Among this plethora of solutions, I decided to focus on meta-learning algorithms, which currently achieve state of the art results in few-shot image classification, in addition to exploiting a conceptually fascinating paradigm. The next section proposes a formulation of this paradigm.

\subsection{Meta-learning paradigm} \label{meta-learning}
Thrun \& Pratt (1998)~\cite{thrun1998learning} stated that, given a task, an algorithm is \emph{learning} \say{if its performance at the task improves with experience}, while, given a family of tasks, an algorithm is \emph{learning to learn} if \say{its performance at each task improves with experience \textbf{and} with the number of tasks}. We will refer to the last one as a \emph{meta-learning algorithm}. Formally, if we want to solve a task $\mathcal{T}_{test}$, the meta-learning algorithm will be trained on a batch of training tasks $\lbrace \mathcal{T}_i \rbrace$. The training experience gained by the algorithm from its attempts at solving these tasks will be used to solve the ultimate task $\mathcal{T}_{test}$.

I will now formalize the meta-learning framework applied to the few-shot classification problem described in section \ref{few-shot-classification}. A visualization is available in Figure \ref{fig:meta-learning}.

\begin{figure}[h]
    \centering
    \includegraphics[width=15cm]{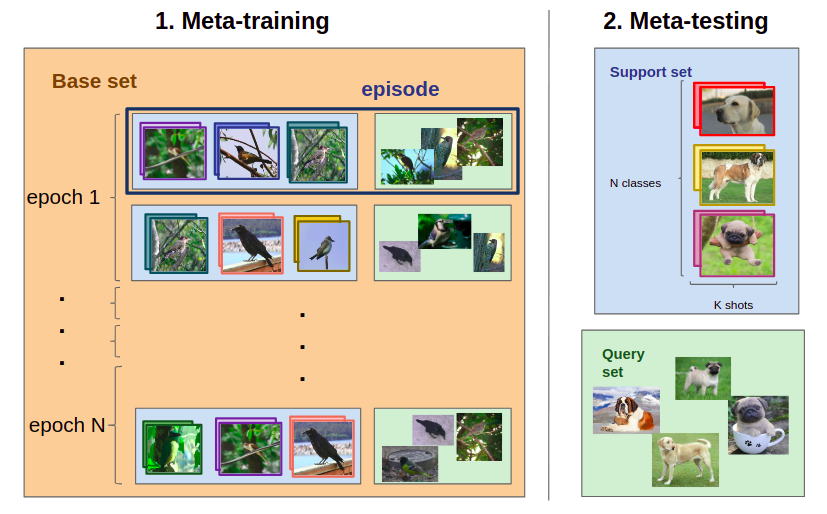}
    \caption{To solve a few-shot image classification task $\mathcal{T}_{test}$ defined by a support set and a query set (on the right), we use a meta-training set $\mathcal{D}$ (on the left) from which we sample episodes in the form of tasks $\mathcal{T}_i$ similar to $\mathcal{T}_{test}$.}
    \label{fig:meta-learning}
\end{figure}

To solve a $N$-way $K$-shot classification problem named $\mathcal{T}_{test}$, we have at our disposal a large meta-training set $\mathcal{D}$. The meta-training procedure will consist of a finite number of episodes.

An episode is composed of a classification task $\mathcal{T}_i$ that is similar to the classification task $\mathcal{T}_{test}$ we ultimately want to solve: from $\mathcal{D}$ we sample $N$ classes and $K$ support-set images for each class, along with $Q$ query images. Note that the classes of $\mathcal{T}_i$ are entirely disjoint from the classes of $\mathcal{T}_{test}$ (\emph{i.e.} the classes of $\mathcal{T}_{test}$  do not appear in the meta-training set $\mathcal{D}$, although they have to be similar for the algorithm to be efficient).
At the end of each episode, the parameters of our model will be trained to maximize the accuracy of the classification of the $Q$ query images (typically by backpropagating a classification loss such as negative log-probability). Thus our model learns across tasks the ability to solve an unseen classification task.

Formally, where a standard learning classification algorithm will learn a mapping $image \mapsto label$, the meta-learning algorithm typically learns a mapping $support \space set \mapsto \left( query \mapsto label \right)$.

The efficiency of our meta-learning algorithm is ultimately measured on its accuracy on the target classification task $\mathcal{T}_{test}$.

\subsection{Meta-learning algorithms}
Recently, several meta-learning algorithms for solving few-shot image classification are published every year. The majority of these algorithm can be labeled as either a metric learning algorithm or as a gradient-based meta-learner. Both kind are presented in this section.

\subsubsection{Gradient-based meta-learning} \label{gradient-based-meta-learning}
In this setting, we distinguish the meta-learner, which is the model that learns across episodes, and a second model, the base-learner, which is instantiated and trained inside an episode by the meta-learner.

Let us consider an episode of meta-training, with a classification task $\mathcal{T}_d$ which is defined by a support set of $N*K$ labeled images and a query set of $Q$ images. The base-learner model, typically a CNN classifier, will be initialized, then trained on the support set (\emph{e.g.} the base-training set). The algorithm used to train the base-learner is defined by the meta-learner model. The base-learner model is then applied to predict the classification of the Q query images. The meta-learner’s parameters are trained at the end of the episode from the loss resulting from the classification error.

\begin{figure}[t]
    \centering
    \includegraphics[width=16.5cm]{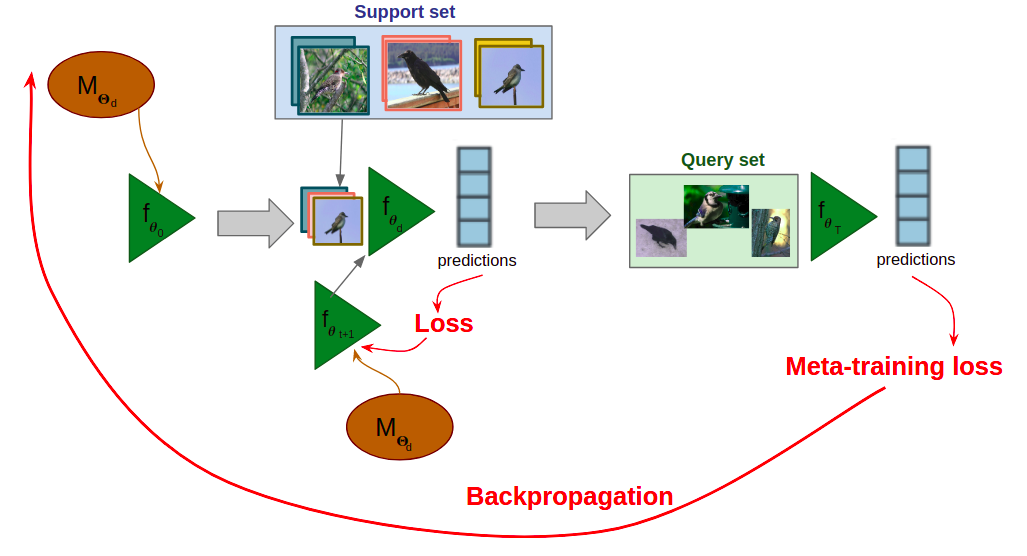}
    \caption{The $d^{th}$ episode of meta-training, which follows this process: $(1)$ the support set and the query set are sampled from the meta-training set; $(2)$ the base-model $f_\theta$ is initialized by the meta-model $M_{\Theta_d}$; $(3)$ the parameters of the base-model are fine-tuned on the support set (the fine-tuning process depends on $M_{\Theta_d}$); $(4)$ after $T$ updates, the base-model is evaluated on the query set; $(5)$ the parameters $\Theta$ of the meta-model are updated by backpropagating the loss resulting from the base-model's predictions on the query set.}
    \label{fig:meta-learning}
\end{figure}

From this point, algorithms differ on their choice of meta-model.

\paragraph{Meta-LSTM (2016)} Ravi \& Larochelle~\cite{Ravi16} decided to use a Long-Short-Term-Memory network~\cite{hochreiter1997long}: the parameters $\theta$ of the base-learner $f_{\theta}$ are represented by the cell state of the LSTM, which leads to the update rule $\theta_t = f_t \odot \theta_{t-1} + i_t \odot \Tilde{c}_t$ where $f_t$ and $i_t$ are respectively the forget gate and the input gate of the LSTM, and $\Tilde{c}_t$ is an input.
We can see the update rule as an extension of the backpropagation, since with $f_t = 1$, $i_t$ the learning rate and $\Tilde{c}_t = - \bigtriangledown_{\theta_{t-1}} \mathcal{L}_t$ we obtain the standard backpropagation.
Hence this model learns how to efficiently operate gradient descents on the base-model from the support set, in order to make this base-model more accurate on the query set.

\begin{figure}[t]
    \centering
    \includegraphics[width=8cm]{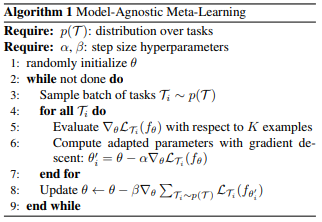}
    \caption{Overview of the MAML algorithm with one gradient update on the support set (credits to~\cite{Finn17})}
    \label{fig:maml-algo}
\end{figure}

\paragraph{Model-Agnostic Meta-Learning (2017)} Finn \emph{et al.}~\cite{Finn17} proposed an algorithm that learns how to initiate the parameters of the base-model, but does not intervene in the base-model's parameters update. Here, the meta-learner creates a copy of itself at the beginning of each episode, and this copy (the base-model) is fine-tuned on the support set, then makes predictions on the query set. The loss computed from these predictions is used to update the parameters of the meta-model (hence, the initialization parameters for the next episodes will be different). The algorithm as described by Finn \emph{et al.} is shown in Figure \ref{fig:maml-algo}.

The main feature of this method is that it is conceived to be agnostic of the base-model, which means that it can virtually be applied to any machine learning algorithm. Finn \emph{et al.} tested it on supervised regression and classification, and on reinforcement learning tasks, but it could be used to solve many other problems necessitating fast adaptation of a Deep Neural Network, for instance for few-shot object detection (see section \ref{YOLOMAML}).

\subsubsection{Metric Learning} \label{metric-learning}
In section \ref{few-shot-classification}, I presented the Siamese Networks algorithm~\cite{Koch15}, which was a first attempt at solving few-shot classification using metric learning, \emph{i.e.} learning a distance function over objects (some algorithms actually learn a similarity function, but they are nonetheless referred to as metric learning algorithms).

As such, metric learning algorithms learn to compare data instances. In the case of few-shot classification, they classify query instances depending on their similarity to support set instances. When dealing with images, most algorithm train a convolutional neural network to output for each image an embedding vector. This embedding is then compared to embeddings of other images to predict a classification.

\paragraph{Matching Networks (2016)} As explained in section \ref{few-shot-classification}, Siamese Networks train their CNN in a discrimination task (\emph{are these two instances from the same class?}) but the algorithm is tested on a classification task (\emph{to which class does this instance belong?}). This issue of task shift between training and testing time is solved by Vinyals \emph{et al.}~\cite{Vinyals16}. They proposed the Matching Networks, which is the first example of a metric learning algorithm inside the meta-learning framework.

To solve a few-shot image classification task, they use a large meta-training set from which they sample episodes (see Figure \ref{fig:meta-learning}). For each episode, they apply the following procedure:

\begin{enumerate}
    \item Each image (support set and query set) is fed to a CNN that outputs as many embeddings;
    \item Each query image is classified using the softmax of the cosine distance from its embedding to the embeddings of support set images;
    \item The cross-entropy loss on the resulting classification is backpropagated through the CNN;
\end{enumerate}

This way, the Matching Networks learn to compute a representation of images that allows to classify them with no specific prior knowledge on the classes, simply by comparing them to a few instances of these classes. Since considered classes are different in every episode, Matching Networks are expected to compute features of the images that are relevant to discriminate between classes, whereas a standard classification learning algorithm is expected to learn the features that are specific to each class.

It is to be noted that Vinyals \emph{et al.} also proposed to augment their algorithm with a Full Context Embedding process: the embedding of each image depends on the embeddings of the others thanks to bidirectional LSTM. They expect that this better exploit all the available knowledge on the episode. This process slightly improved the performance of their algorithm on the miniImageNet benchmark, but also demands a longer computing time.

\paragraph{Prototypical Networks (2017)} Building on Matching Networks, Snell \emph{et al.}~\cite{Snell17} proposed Prototypical Networks. The process is essentially the same (although Full Context Embeddings are not used), but a query image is not compared to the embeddings of every images of the support set. Instead, the embeddings of the support set images that are from the same class are averaged to form a class prototype. The query image is then compared only to these prototypes. It is to be noted that when we only have one example per class in the support set (One-Shot Learning setting) the Prototypical Networks are equivalent to the Matching Networks.
They obtained better results than Matching Networks on the miniImageNet benchmark, and expose that part of this improvement must be credited to their choice of distance metric: they notice that their algorithm and Matching Networks both perform better using Euclidean distance than when using cosine distance.

\begin{figure}[t!]
    \centering
    \includegraphics[width=12cm]{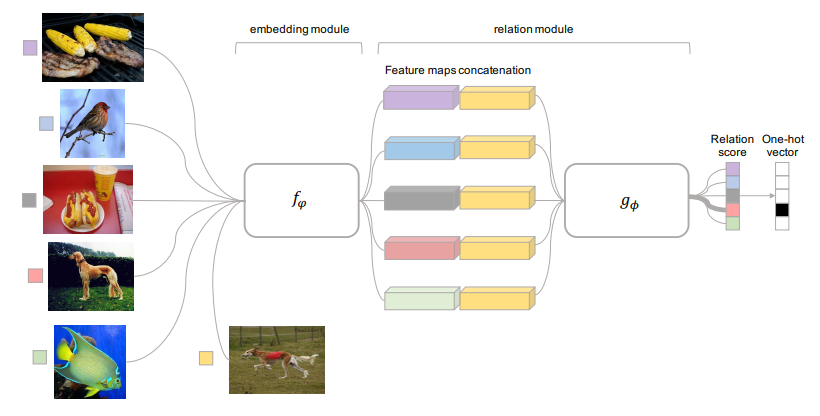}
    \caption{Relation Network architecture for a 5-way 1-shot problem with one query example (credits to~\cite{Sung18}). Note that they chose to represent the final output with a one-hot vector obtained by a \emph{max} function on the relation scores, but that during training time we need to use a \emph{softmax} to make the network differentiable.}
    \label{fig:relation-network}
\end{figure}

\paragraph{Relation Network (2018)} Sung \emph{et al.}~\cite{Sung18} built on Prototypical Networks to develop the Relation Network. The difference is that the distance function on embeddings is no longer arbitrarily defined in advance, but learned by the algorithm (see Figure \ref{fig:relation-network}): a \emph{relation module} is put on top of the \emph{embedding module} (which is the part that computes embeddings and class protoypes from the input images). This relation module is fed the concatenation of the embedding of a query image with each class prototype, and for each couple outputs a relation score. Applying a softmax to the relation scores, we obtain a prediction.

\subsection{Few-Shot Image classification benchmarks} \label{datasets}

Algorithms intended to solve the few-shot learning problem are usually tested on two datasets: Omniglot and miniImageNet.

\begin{figure}[h]
    \centering
    \includegraphics[width=5cm]{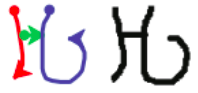}
    \caption{Two different visualizations of a same instance of the Omniglot dataset. On the left, we can see how the character was drawn. On the right, we see a 28x28 one-channel image. Credits to~\cite{lake2015human}}
    \label{fig:omniglot-instance}
\end{figure}

\paragraph{Omniglot} Lake \emph{et al.} (2011)~\cite{lake2011} introduced the Omniglot dataset. It is composed of 1623 characters from 50 distinct alphabets. Each one of these characters is a class and contains 20 samples drawned by distinct people. Each data instance is not only a 28x28x1 image, but also contains information about how it was drawn: how many strokes, and the starting and ending point of each stroke (see Figure \ref{fig:omniglot-instance}). Although Lake \emph{et al.} primarily used Omniglot for few-shot learning of visual concepts from their subparts~\cite{lake2015human}, the dataset as a set of 28x28 one-channel images is used as a MNIST-like benchmark for few-shot image classification. Most algorithm now achieve a 98\%-or-better accuracy on this dataset on most use cases~\cite{Sung18}.

\paragraph{miniImageNet} Vinyals \emph{et al.}~\cite{Vinyals16} proposed to use a part of ImageNet as a new, more challenging benchmark for few-shot image classification. Their dataset consist of 100 classes, each containing 600 3-channel images. The commonly used train/validation/evaluation split of this dataset~\cite{Ravi16} separates it in three subsets of respectively 64, 16 and 20 classes. This way, we ensure that the algorithm is evaluated on classes that were not seen during training.

\subsection{Few-Shot object detection}
Although research in few-shot object detection is currently less advanced than in few-shot classification, some solutions to this problem have been proposed in the last few months. First, we will go over the existing solutions for standard object detection, then we will learn about the recent efforts in developing algorithms for few-shot object detection.

\paragraph{Object detection} Algorithms for object detection can be separated in two categories: single-stage detectors and the R-CNN family (two-stage detectors). Single-stage detectors aim at performing fast detection while algorithms like R-CNN are more accurate.

R-CNN~\cite{girshick14} uses a first network to determine regions of interest in an image, and then a second network to classify the content of each region of interest. Fast R-CNN~\cite{girshick2015fast} and Faster R-CNN~\cite{ren2015faster} improved the algorithm's efficiency by reducing redundant computations and the number of regions of interest. Mask R-CNN~\cite{he2017mask} uses the same principle as R-CNN but performs image segmentation.

Single-stage detectors perform object detection on an image in a single forward pass through a CNN: the bounding-box and the label of each object are predicted concurrently. Leading single-stage detectors are the SSD (for Single-Shot Detector)~\cite{liu2016ssd}, RetinaNet~\cite{lin2017focal} and YOLO (for You Only Look Once)~\cite{redmon2016you}.

\begin{figure}[t]
    \centering
    \includegraphics[width=16cm]{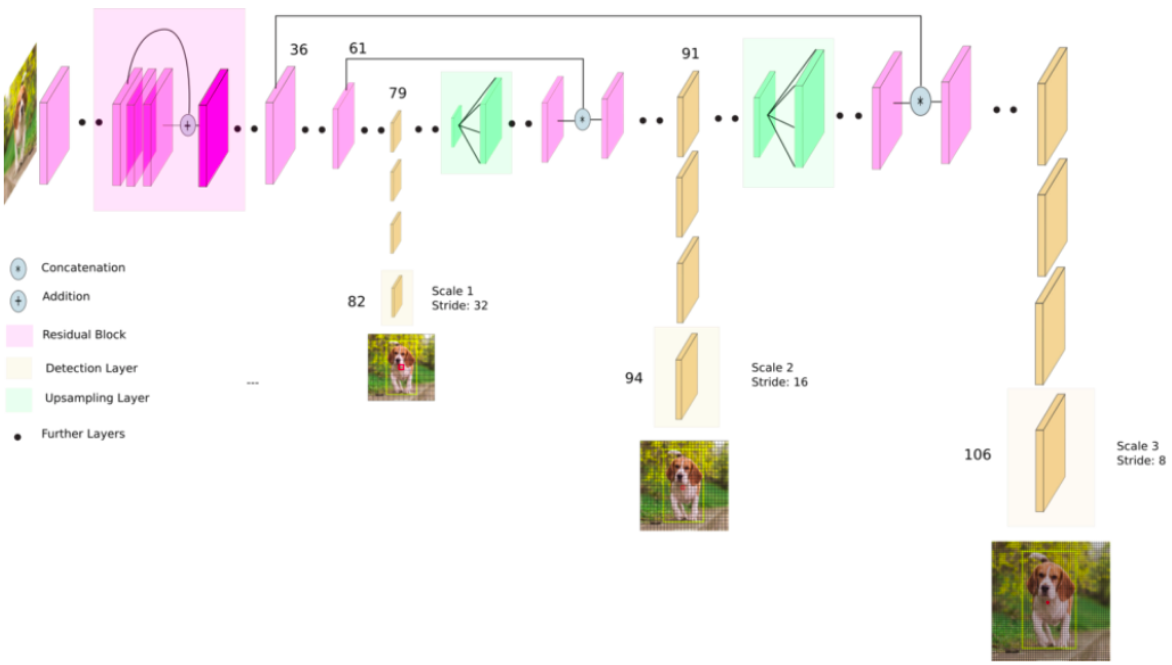}
    \caption{Architecture of YOLOv3. Credits to Ayoosh Kathuria.}
    \label{fig:yolov3}
\end{figure}

YOLO went through two incremental improvements since its creation in 2016. Its last version, YOLOv3, contains three output layers. Each one is responsible for predicting respectively large, medium-size and small objects. For each output layer, three \emph{anchors} are set as hyperparameters of the model. An anchor is like a "default bounding box", and YOLOv3 actually predicts deformations to these anchors, rather than predicting a bounding box from scratch. The network is mostly composed of residual blocks~\cite{he2016deep}. In particular, the backbone of the model is a Darknet53, a 53-layer residual network pre-trained on ImageNet. A visualization of the YOLOv3 architecture is available in Figure \ref{fig:yolov3}.

\paragraph{Few-Shot Object Detectors} To the best of my knowledge, the first few-shot detector was proposed in late 2018 by Kang \emph{et al.}~\cite{kang2018few}. Their algorithm combines a standard single-stage detector with an other auxiliary network. This second model is responsible for reweighting the features outputted by the feature extractor of the model (in YOLOv3, this would be the output of the Darknet53). The goal of this reweighting is to give more importance to features related to the specific few-shot detection task being solved (the intuition is that the relevant features for detection depends of the type of object to detect). The reweighting model is trained in a meta-learning set-up (see section \ref{meta-learning}): at each meta-training episode, a few-shot detection task is sampled (for instance: detecting dogs and cats, with a few annotated examples of dogs and cats), and the training objective is the precision of the detector.

More recently, Fu \emph{et al.} proposed the Meta-SSD~\cite{fu2019meta}. They apply Li \emph{et al.}'s Meta-SGD~\cite{li2017meta} to Liu \emph{et al.}'s Single-Shot Detector. They end up with a fully meta-trainable object detector.

Concurrently, Wang \emph{et al.}~\cite{wang19} developed a novel framework around the Faster R-CNN. The resulting algorithm can adapt to a new task with few labeled examples.

Previous works already tackled few-shot object detection~\cite{chen2018lstd}~\cite{Dong2017FewshotOD}, although they considered a slightly different problem: they defined few-shot as few labeled images per category, but also used a large pool of unlabeled data. 

\newpage
\section{Contributions}

\subsection{Overview}
Sicara is a company which develops computer vision solutions based on machine learning algorithms for businesses. However, it is common that the amount of data made available by the customer is not large enough to effectively train a standard convolutional neural network. Also, we often need to harness this data with a very short lead time. Therefore, a company like Sicara needs an efficient and ready-to-use meta-learning algorithm for few shot learning problems related to computer vision.

I was in charge of the first step of this process, which is benchmarking several state-of-the-art algorithms, identifying the strengths and weaknesses of each algorithm, its performance on different kinds of datasets, and overall their relevance depending on the task that needs solving.

During this work on meta-learning algorithms, we decided to focus on the Model Agnostic Meta-Learner~\cite{Finn17} and to switch from the few-shot image classification problem to the few-shot object detection problem, which had until then attracted less attention in the research community than few-shot classification. Our idea is to apply MAML to the YOLOv3 object detector in order to obtain an algorithm capable of detecting new classes of objects with little time and only a few examples.

In this section, I will first explain my work on meta-learning algorithms for few-shot image classification, then I will detail my progress so far in developing a novel algorithm: the YOLOMAML.

\subsection{Meta-learning algorithms for Few-Shot image classification}
I compared the performance of four meta-learning algorithms, using two datasets: miniImageNet (see section \ref{datasets}) and Caltech-UCSD Birds 200 (CUB)~\cite{WelinderEtal2010}, which is the dataset containing 6,033 pictures of birds from 200 different classes. The four algorithms are the following:
\begin{itemize}
    \item Matching Networks~\cite{Vinyals16}
    \item Prototypical Networks~\cite{Snell17}
    \item Relation Network~\cite{Sung18}
    \item Model Agnostic Meta-Learner~\cite{Finn17}
\end{itemize}

The primary intention was to conduct extensive experiments on these algorithms with variations on both their settings, the target tasks and the training strategy, in order to obtain a fine understanding of how these algorithms behave and how to best harness their abilities. I also intended to include other promising algorithms, such as the Simple Neural Attentive Learner~\cite{Mishra18} or the Meta-LSTM~\cite{Ravi16}. However, since we decided halfway through the benchmark to focus on the exciting opportunity of developing a novel meta-learning object detector, there wasn't enough time to go through the all set of experiments. Hence, my contribution for a deeper understanding of meta-learning consists in:
\begin{enumerate}
    \item a documented implementation of meta-learning algorithms for few-shot classification tasks, with a focus on allowing future researchers in the field to easily launch new experiments, in a clear and reproducible way;
    \item the reproduction of the results presented by Chen \emph{et al.}~\cite{Chen19}, bringing the exposition of the challenges that we face when benchmarking meta-learning algorithms;
    \item a study on the impact of label noise in the support set at evaluation time;
\end{enumerate}

In this subpart I will present these contributions with more details, both on the results and on the process of obtaining these results.

\subsubsection{Implementation}
Chen \emph{et al.}~\cite{Chen19} published in April 2019 a first unified comparison of meta-learning algorithms for few-shot image classification, and made their source code available\footnote{\url{https://github.com/wyharveychen/CloserLookFewShot}}. For us, this code in \texttt{PyTorch} presents two main advantages:
\begin{enumerate}
    \item It proposes a unified implementation of Matching Networks, Prototypical Networks, Relation Network, MAML and two baseline methods for comparison. This allows the experimenter to fairly compare algorithms.
    \item It contains a relatively consistent framework for the treatment of the several datasets (Omniglot, EMNIST~\cite{cohen2017emnist}, miniImageNet and CUB), from the description of the train / validation / evaluation split using \texttt{json} to the sampling of the data in the form of episodes for few-shot image classification, which uses the code from Hariharan \emph{et al.}~\cite{hariharan2017low}\footnote{\url{https://github.com/facebookresearch/low-shot-shrink-hallucinate}}.
\end{enumerate}

For these reasons, I used this code as a (very advanced) starting point for my implementation. I identified three main issues:
\begin{enumerate}
    \item The original code was very scarcely documented, which makes it difficult to understand, and even more difficult to modify, since it was not always clear what a chunk of code did, or what a variable represented.
    \item Some experiment parameters were defined inside the code and therefore not easily customizable when launching an experiment, nor monitorable after the experiments, affecting the reproducibility of the experiments.
    \item Some chunks of code were duplicated in several places in the project.
\end{enumerate}

The main goal of my work on this code was to make it easily accessible, allowing future researcher to understand the way these algorithms work in practice, and to quickly be able to launch their own experiments. This goal was achieved by:
\begin{itemize}
    \item cleaning the code and removing all duplicates;
    \item extensively document every class and function with the knowledge gained during my work on the code;
    \item integrate two useful tools for conducting experiments:
    \begin{itemize}
        \item \texttt{pipeline} is an internal library at Sicara which allows to configure experiments with a YAML file: this file describes the different steps of the experiment and explicitly indicates all necessary parameters of the experiment;
        \item \texttt{Polyaxon} is an open-source platform or conducting machine learning experiments; its main features (for our usage) are $(1)$ an intuitive dashboard for keeping track of all passed, current and programmed experiments, with for each one the YAML configuration file, along with all logs and ouputs, $(2)$ the possibility to launch groups of experiments with varying parameters, and $(3)$ a Tensorboard integrated to the platform.
    \end{itemize}
\end{itemize}

\begin{figure}[h]
    \centering
    \includegraphics[width=8cm]{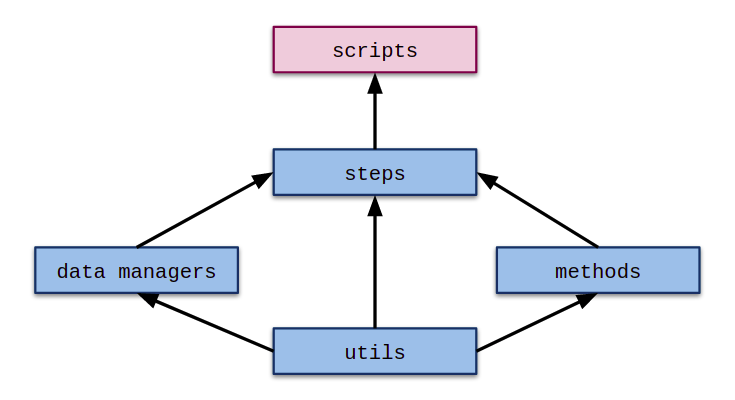}
    \caption{Structure of my code to conduct experiments on meta-learning algorithms for few-shot image classification}
    \label{fig:code-structure}
\end{figure}

The structure of the implementation is shown in Figure \ref{fig:code-structure}. The code can be divided in five categories, detailed below. \footnote{The code is available at \url{https://github.com/ebennequin/FewShotVision}. Note that this version does not use the \texttt{pipeline}, which is private to Sicara.}

\paragraph{\texttt{scripts}} These are the files that must be executed to launch the experiments. I used YAML files for compatibility with the \texttt{pipeline} library, but standard Python scripts could be used just as well (and are actually used in the publicly available repository). They describe the flow between the different steps (which steps uses which step outputs) and contain all the values parameterizing the experiment:
\begin{itemize}
    \item dataset to work on (ex: miniImageNet);
    \item method to work with (ex: Matching Networks);
    \item backbone CNN of the method (ex: Resnet18);
    \item parameters of the $N$-way $k$-shot classification task with $q$ queries per class (with $N$ allowed to be different at training and evaluation time);
    \item whether to perform data augmentation on the meta-training set;
    \item number of meta-training epochs;
    \item number of episodes (\emph{i.e.} classification tasks) per meta-training epoch;
    \item optimizer (ex: Adam);
    \item learning rate;
    \item which state of the model to keep for evaluation (the model trained on all the epochs, or the model that achieve the best validation accuracy);
    \item number of few-shot classification task to evaluate the model on;
\end{itemize}

\paragraph{\texttt{steps}} They are called by the scripts and use the parameters explicited in it. One example of step is \texttt{MethodTraining}, which is responsible for the training of the model.

\paragraph{\texttt{data managers}} They define the \texttt{SetDataset} and \texttt{EpisodicBatchSampler} classes, which respectively extend the \texttt{PyTorch} base classes \texttt{Dataset} and \texttt{Sampler} and are used to build a \texttt{DataLoader} that loads the data in the shape of few-shot classification task (\emph{i.e.} a support set and a query set, instead of regular batches of arbitrary size).

\paragraph{\texttt{methods}} Each file in this category defines a class corresponding to one meta-learning algorithm (ex: Prototypical Networks). Every class contains three essential methods:
\begin{itemize}
    \item \texttt{set\_forward(episode)}: takes as input an episode composed of a support set and a query set, and outputs the predictions of the model for the query set.
    \item \texttt{train\_loop()}: executes one meta-training epoch on the meta-training set.
    \item \texttt{eval\_loop()}: evaluates the model on few-shot classification tasks sampled from the evaluation set.
\end{itemize}

\paragraph{\texttt{utils}} These files contain all the utilities used in the rest of the code.


\subsubsection{Reproducing the results} \label{reproduction}
The first thing to do with this reimplementation was to validate it by reproducing the results reported by Chen \emph{et al.}~\cite{Chen19}. This unexpectedly granted us with interesting new knowledge.

\begin{figure}[h]
    \centering
    \includegraphics[width=15cm]{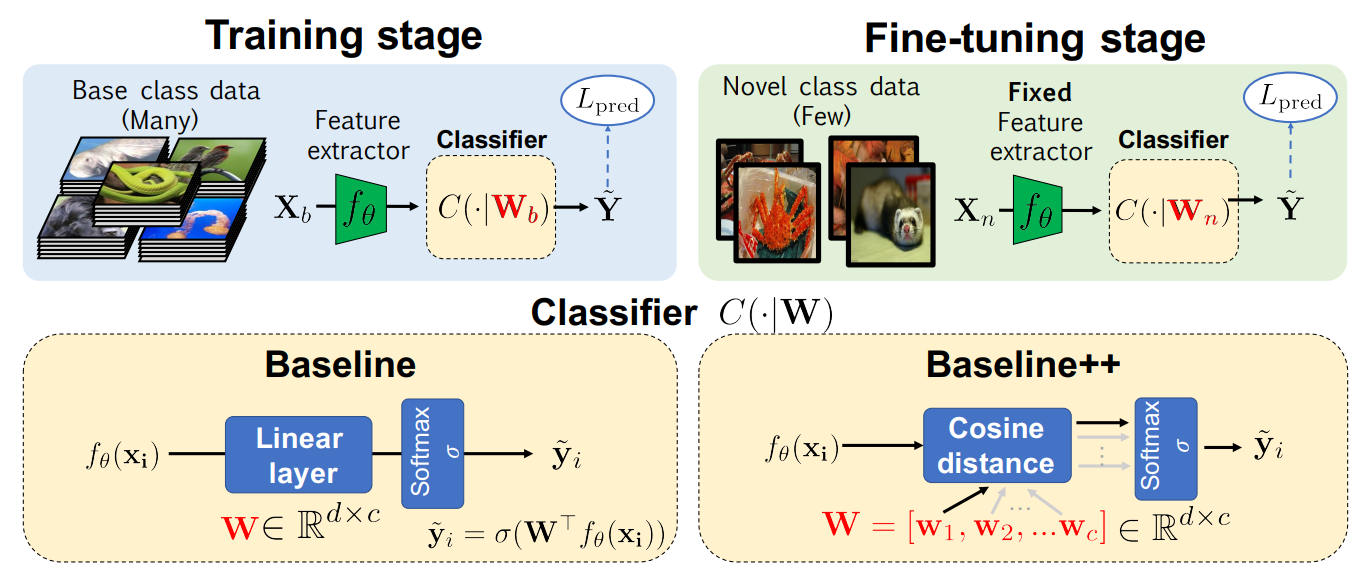}
    \caption{Baseline and Baseline++ few-shot classification methods. Both algorithms are pre-trained on the meta-training set. When evaluated on a few-shot classification task, the feature extractor $f_\theta$ is freezed and the classifier $\mathcal{C}$ is fine-tuned on the support set before being applied to the query set. In Baseline++, the classifier is not a standard fully connected layer, but computes the cosine distance between its weights and the input features vector. Both algorithm are used to compare the meta-learning algorithms to non-meta-learning methods. This figure is credited to~\cite{Chen19}.}
    \label{fig:baselines}
\end{figure}

I experimented on the CUB dataset for a shorter running time. I reproduced Chen \emph{et al.}'s experiments in the 5-way 1-shot and 5-way 1-shot settings, for Matching Networks, Prototypical Networks, Baseline and Baseline++ (see Figure \ref{fig:baselines}). I purposefully omitted MAML for this part, since this algorithm's training takes about five times longer than the others' (see Table \ref{tab:running-times}). Relation Network is also omitted because its process is essentially similar to Prototypical Networks.

\begin{table}[h]
    \centering
    \begin{tabular}{|c|c|c|c|c|}
        \hline
        & \multicolumn{2}{c|}{ CUB } & \multicolumn{2}{c|}{ miniImageNet } \\ \hline
        & 1 shot & 5 shots & 1 shot & 5 shot \\ \hline
        Baseline & 1h10 & 1h00 & 10h05 & 10h07 \\ \hline
        Baseline++ & 56mn & 51mn & 10h25 & 10h28 \\ \hline
        MatchingNet & 6h41 & 4h21 & 7h51 & 6h23 \\ \hline
        ProtoNet & 6h38 & 5h07 & 7h40 & 6h08 \\ \hline
        MAML & 28h05 & 22h28 & 31h22 & 25h22 \\ \hline
    \end{tabular}
    \caption{Running time of several algorithms depending on the setting and dataset. This is the running time of the whole process, from training to evaluation.}
    \label{tab:running-times}
\end{table}

The parameters of the experiments follow those described by Chen et al, \emph{i.e.} a 4-layer CNN as a backbone, an Adam optimizer with a learning rate of $10^{-3}$, 100 episodes per epoch and data augmentation on the training set. The baselines are trained for 200 epochs on CUB, an for 400 epochs on miniImageNet. The other algorithms are trained for 600 epochs in the 1-shot setting, and on 400 epochs in the 5-shots setting. We keep the state of the model that had the best accuracy on the validation set, and evaluate it on 600 few-shot classification tasks sampled from the evaluation set. 

The results of these experiments are reported in Table \ref{tab:reprod1}. 6 out of 8 experiments gave results out of the 95\% confidence interval reported by Chen et al, with a difference up to 6\% in the case of 1-shot Baseline++. Our results fall below the confidence interval in 4 cases and above the confidence interval in 2 cases.

\begin{table}[h]
    \centering
        \begin{tabular}{|c|c|c|c|c|}
        \hline
        & \multicolumn{2}{c|}{ our reimplementation } & \multicolumn{2}{c|}{ Chen \emph{et al.}'s } \\ \hline
        & 1 shot & 5 shots & 1 shot & 5 shot \\ \hline
        Baseline & 46.57 $\pm$ 0.73 & \textbf{68.36 $\pm$ 0.66} & 47.12 $\pm$ 0.74 & 64.16 $\pm$ 0.71 \\ \hline
        Baseline++ & \textbf{53.71 $\pm$ 0.82} & \textbf{75.09 $\pm$ 0.62} & 60.53 $\pm$ 0.83 & 79.34 $\pm$ 0.61 \\ \hline
        MatchingNet & \textbf{58.43 $\pm$ 0.85} & \textbf{75.52 $\pm$ 0.71} & 61.16$\pm$ 0.89 & 72.86 $\pm$ 0.70 \\ \hline
        ProtoNet & 50.96 $\pm$ 0.90 & \textbf{75.48 $\pm$ 0.69} & 51.31 $\pm$ 0.91 & 70.77 $\pm$ 0.69 \\ \hline
        \end{tabular}
    \caption{Comparison of the results of our reimplementation compared to the results reported by Chen \emph{et al.}, on the CUB dataset with a 5-way classification task. Our results are shown in bold when they are out of the 95\% confidence interval reported by Chen \emph{et al.}}
    \label{tab:reprod1}
\end{table}

A fair assumption was that my implementation was to blame for this incapacity to reproduce the original paper's results. To verify it, I reproduced the experiments of Chen \emph{et al.} using their original implementation. The results are shown in Table \ref{tab:reprod2}. In most cases, they are out of the 95\% confidence interval reported in~\cite{Chen19}.

\begin{table}[h]
    \centering
    \begin{tabular}{|c|c|c|c|c|}
        \hline
        & \multicolumn{2}{c|}{ CUB } & \multicolumn{2}{c|}{ miniImageNet } \\ \hline
        & 1 shot & 5 shots & 1 shot & 5 shot \\ \hline
        Baseline++ & 61.36 $\pm$ 0.92 & \textbf{77.53 $\pm$ 0.64} & 48.05 $\pm$ 0.76 & 67.01 $\pm$ 0.67 \\ \hline
        MatchingNet & \textbf{59.55 $\pm$ 0.89} & \textbf{75.63 $\pm$ 0.72} & 48.43 $\pm$ 0.77 & \textbf{62.26 $\pm$ 0.70} \\ \hline
        ProtoNet & \textbf{50.28 $\pm$ 0.90} & \textbf{75.83 $\pm$ 0.67} & 43.89 $\pm$ 0.73 & \textbf{65.55 $\pm$ 0.73} \\ \hline
        MAML & \textbf{54.57 $\pm$ 0.99} & \textbf{75.51 $\pm$ 0.73} & \textbf{43.92 $\pm$ 0.77} & 62.96 $\pm$ 0.72 \\ \hline
    \end{tabular}
    \caption{Reproduction of the results of~\cite{Chen19} on both CUB and miniImageNet, using the implementation provided with the paper. Our results are shown in bold when they are out of the 95\% confidence interval reported in~\cite{Chen19}.}
    \label{tab:reprod2}
\end{table}

From there, my assumption was that the incertitude on the results didn't come solely from the sampling of the evaluation tasks, but also from the training. I proceeded to verify this assumption. I relaunched the first experiment 8 times for Prototypical Networks and evaluated the 8 resulting model on the exact same classification tasks. The results are shown in Table \ref{tab:reprod3}. We can see that the accuracy can go from 74.20\% to 76.04\% on the same set of tasks. This validates that two same models trained with the same hyperparameters may obtain different accuracies on the same evaluation tasks.

\begin{table}[h]
    \centering
    \begin{tabular}{|c|}
        \hline
        76.04 $\pm$ 0.71 \\ \hline
        74.77 $\pm$ 0.71 \\ \hline
        75.45 $\pm$ 0.71 \\ \hline
        75.54 $\pm$ 0.70 \\ \hline
        75.14 $\pm$ 0.70 \\ \hline
        74.90 $\pm$ 0.71 \\ \hline
        74.91 $\pm$ 0.71 \\ \hline
        74.20 $\pm$ 0.70 \\ \hline
    \end{tabular}
    \caption{Accuracy of the Prototypical Network on the same set of evaluation tasks on the CUB dataset in the 5-way 5-shot setting, after 8 independent training processes.}
    \label{tab:reprod3}
\end{table}

From this work on the reproduction of the results reported by Chen \emph{et al.}, we can retain two main take-aways:

\begin{enumerate}
    \item The results obtained with a instance of \emph{meta-training + evaluation} cannot be reproduced, although the \texttt{numpy} random seed is systematically set to 10 at the beginning of the process. I learned that setting the \texttt{numpy} random seed is not enough to fix the randomness of a training using \texttt{PyTorch}. I found that the following lines succeed in doing so:
    \begin{lstlisting}[language=Python]
    np.random.seed(numpy_random_seed)
    torch.manual_seed(torch_random_seed)
    torch.backends.cudnn.deterministic = True
    torch.backends.cudnn.benchmark = False
    \end{lstlisting}
    The third and fourth lines are only necessary when using the CuDNN backend\footnote{\url{https://pytorch.org/docs/stable/notes/randomness.html}}.
    \item On the same set of evaluation tasks, the accuracy of a model can vary with an amplitude of up to 2\% due to ramdomness in the training. This amplitude is similar to the reported difference in accuracy between algorithms and higher than the confidence intervals usually reported when evaluating meta-learning algorithms~\cite{Chen19}~\cite{Sung18}~\cite{Finn17}~\cite{Ravi16}~\cite{Snell17}~\cite{Bertinetto18}. I argue that a reported difference of a few percents in accuracy between two meta-learning algorithms on a set of classification tasks cannot be considered as a relevant comparator of these algorithms. It would be ideal to get an exact measure of the uncertainty by launching a sufficient number of trainings, but the necessary computing time for this operation is prohibitive (see Table \ref{tab:running-times}).
\end{enumerate}

\subsubsection{Effects of label noise in the support set at evaluation time}
In practice, meta-learning algorithms can be used this way:
\begin{enumerate}
    \item The model is trained once and for all by the model's designer on a large dataset (with a possibility to update when new labeled examples become available);
    \item When faced with a novel few-shot classification task, the user feeds a few labeled examples to the model, and then is able to apply it to the query images.
\end{enumerate}
As the model's designer and the model's user can be different entities, and as the source of the support set for the novel task may be different from the source of the meta-training data, the designer may not be able to control the quality of the data in the novel task. This is why the model's robustness to noisy data in the support set is an important issue.

In this subsection, we address the issue of label noise (\emph{i.e.} data instances assigned with the wrong label) in the support set of the evaluation classification task. To simulate this noise, we use label swaps: given an integer $M$, for each classification task, we execute $M$ label swaps on the support set of the classification task. Here is the process of one label swap:
\begin{enumerate}
    \item Sample uniformly at random two labels $l_1, l_2$ among the $N$ labels of the support set
    \item For each label $l_x$, select uniformly at random one image $i_{l_x}$ among the $K$ images in the support set associated with this label
    \item Assign label $l_1$ to image $i_{l_2}$ and label $l_2$ to image $i_{l_1}$
\end{enumerate}
Note that even though one label swap changes the label of two images, $M$ label swaps do not necessarily cause $2M$ falsely labeled images, since swaped images are sampled with replacement (in the following, you will see that most models reach an accuracy of 35\% even after 10 label swaps were applied on the \emph{5 labels $\times$ 5 images} support set, which would be hard to explain if 80\% of the support set had false labels).

Also, label swaps are not a perfect simulation: in real cases, the fact that an image supposed to have the label $l_1$ was falsely labeled with $l_2$ does not mean that an other image supposed to have the label $l_2$ was falsely labeled with $l_1$. However, this solution ensures that the support set is still balanced even after applying the artificial label noise (in a 25-images dataset, if one label has one example less than an other label, the dataset becomes noticeably unbalanced). Therefore, we know that the results will not come from errors in labelisation, and not from an unbalanced dataset.

\paragraph{Measuring the effects of label noise in the support set at evaluation time} First, we need to measure the effect of label noise on the model's accuracy. I experimented both on CUB and miniImageNet, with the algorithms Baseline, Baseline++, Matching Networks, Prototypical Networks, Relation Network and MAML. All models were trained on 400 epochs, with the Adam optimizer and a learning rate of $10^-3$. Meta-learning algorithms (\emph{i.e.} all but Baseline and Baseline++) were trained on 5-way 5-shot classification tasks. No artificial label noise was added to the training set.

The models were then evaluated on 5-way 5-shot classification tasks on four different settings corresponding to four different number of label swaps in each classification task (0, 3, 6 and 10). I reported for each setting the mean of the accuracy on 600 tasks. Note that all models (here and in the following of this subsection) are evaluated on the same tasks. To be consistent with my remarks in section \ref{reproduction}, the results are reported with a precision of 1\%.

\begin{figure}[t]
  \begin{minipage}[b]{0.5\linewidth}
   \centering
   \includegraphics[width=8cm]{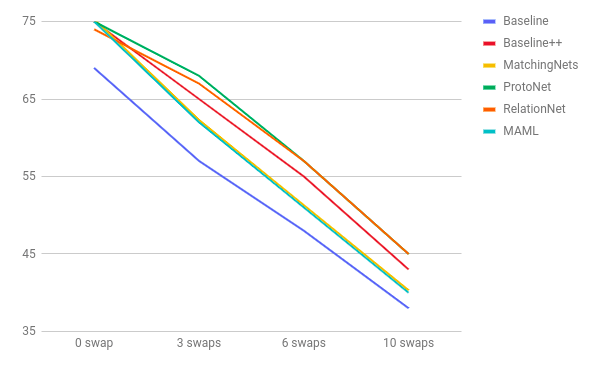}      
  \end{minipage}
\hfill
  \begin{minipage}[b]{0.48\linewidth}
   \centering
   \includegraphics[width=8cm]{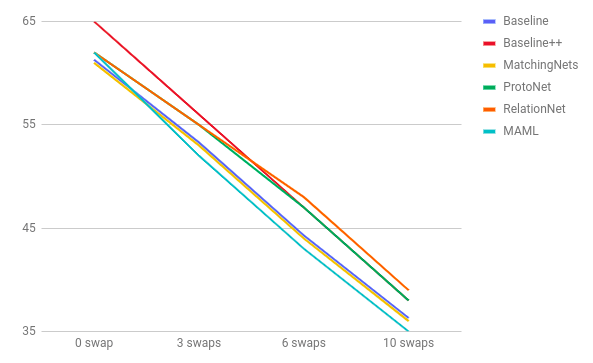}      
  \end{minipage}
  \caption{Accuracy of the methods for different number of label swaps in the support set of each classification task. Left: CUB. Right: miniImageNet.}
  \label{fig:noise-1}

\end{figure}

The results are shown in Figure \ref{fig:noise-1}. We observe that all algorithms endure a serious drop in accuracy on the query set when the label noise in the support set increases, which was expected. We notice that Prototypical Networks and Relation Network are slightly less impacted. This could be explained by the fact that both algorithms use the mean of the features vectors for each class, which reduces the impact of extreme values. 

\paragraph{10-way training} Snell \emph{et al.}~\cite{Snell17} showed that, when evaluating metric learning algorithms on $N$-way $K$-shot classification tasks, the models trained on $N'$-way $K$-shot classification tasks with $N'>N$ performed better than the models trained on $N$-way $K$-shot classification tasks (the intuition being that a model trained on more difficult tasks will generalize better to new tasks, or, in French, \emph{\say{qui peut le plus peut le moins}}). I tested whether this trick also made the model more robust to label noise.

I conducted the same experiment as the one described in the previous paragraph, with the exception that the training was done on 10-way 5-shot classification tasks (instead of 5-way 5-shot). This experiment was done only on metric learning algorithms (\emph{i.e.} Matching Networks, Prototypical Networks, Relation Networks). Indeed, MAML does not allow to change the number of labels in the classification tasks, since the architecture of the CNN (ending in a $N$-filter linear layer) needs to stay the same.

\begin{figure}[t]
  \begin{minipage}[b]{0.5\linewidth}
   \centering
   \includegraphics[width=8cm]{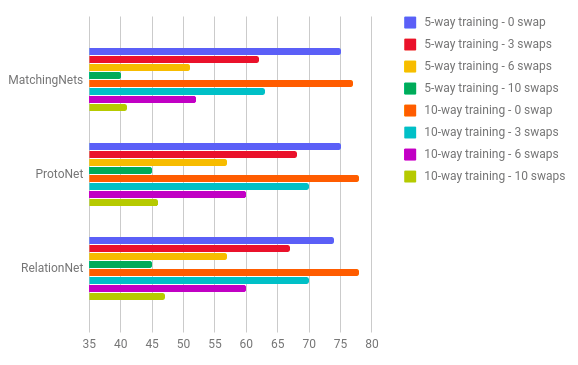}      
  \end{minipage}
\hfill
  \begin{minipage}[b]{0.48\linewidth}
   \centering
   \includegraphics[width=8cm]{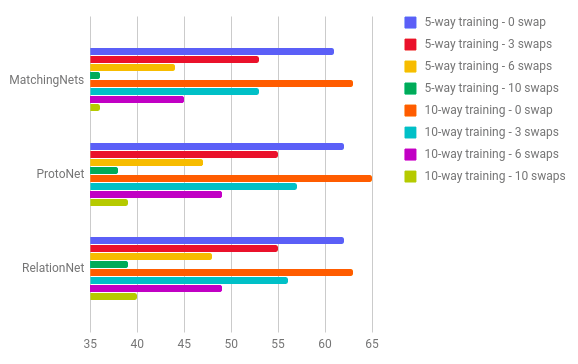}      
  \end{minipage}
  \caption{Accuracy of the methods for different number of label swaps in the support set of each classification task, with a 5-way training and a 10-way training. Left: CUB. Right: miniImageNet.}
  \label{fig:noise-2}

\end{figure}

The results are shown in Figure \ref{fig:noise-2}. They confirm that using a higher number of labels per classification task during training increases the accuracy of the model. However, this doesn't seem to have any effect on the robustness to label noise.

\begin{figure}[t]
  \begin{minipage}[b]{0.5\linewidth}
   \centering
   \includegraphics[width=8cm]{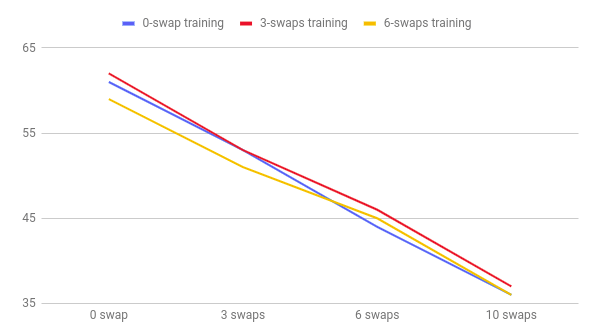}      
  \end{minipage}
\hfill
  \begin{minipage}[b]{0.48\linewidth}
   \centering
   \includegraphics[width=8cm]{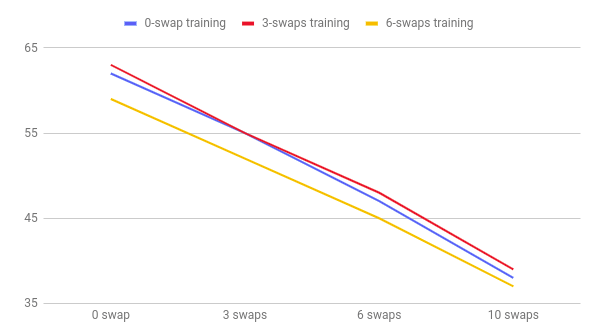}      
  \end{minipage}
  \begin{minipage}[b]{0.5\linewidth}
   \centering
   \includegraphics[width=8cm]{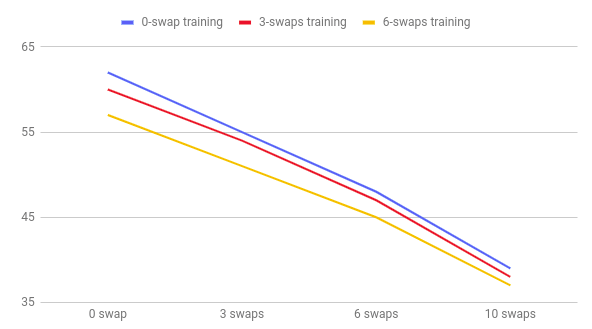}      
  \end{minipage}
\hfill
  \begin{minipage}[b]{0.48\linewidth}
   \centering
   \includegraphics[width=8cm]{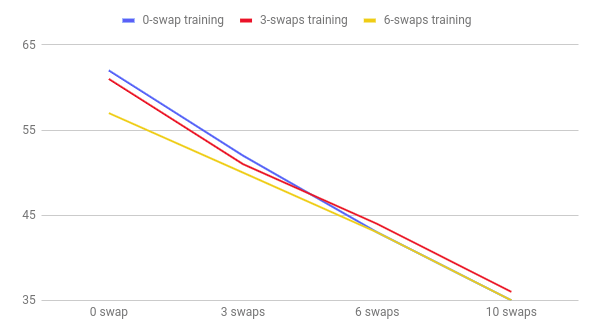}      
  \end{minipage}

  \caption{From left to right, top to bottom: Matching Networks, Prototypical Networks, Relation Network, MAML. For each method, accuracy on a model trained with three strategies, for different number of label swaps in the support set at evaluation time.}
  \label{fig:noise-3}

\end{figure}

\paragraph{Simulating label noise during meta-training} Coming from the idea that training and testing conditions must match, I assumed that incorporating artificial label noise in the support set of the classification tasks on which the models are meta-trained could increase their robustness to label noise at evaluation time. The following experiment tests this assumption.
Label swaps are introduced in the classification tasks composing the meta-training, in the same way that they were applied to the classification tasks at evaluation time in the previous experiments. This results in three set-ups, respectively referred to as 0, 3 and 10-swap training:
\begin{enumerate}
    \item Same experiment as in the first paragraph of this section, only on miniImageNet (not CUB)
    \item Same, but in each episode of the meta-training, 3 label swaps are applied to the support set
    \item Same, but in each episode of the meta-training, 10 label swaps are applied to the support set
\end{enumerate}
Note that we do not experiment on the baselines, since they are not meta-learning algorithm and thus do not solve classification task during training.
The results of this experiment are shown in Figure \ref{fig:noise-3}. We see that adding label swaps during meta-training causes a drop in accuracy when the model is evaluated on correctly labeled classification tasks. The difference is less obvious when the number of label swaps in evaluation tasks increases. Based on these experiments, there is no reason to believe that introducing artificial label noise during the meta-training makes meta-learning algorithms more robust to label noise in novel classification tasks.

\subsubsection{Future work}
In addition to the choice of the meta-learning algorithm, there are many possible ways to improve its performance with minor design choices, such as hyperparameter tuning, or, in the case of Prototypical Networks and their derivatives, the choice of the prototype function. The mean function could be replaced for instance by a "leaky" median (in a way that leaves the function differentiable).

However, we saw that these algorithms only differ by a small margin. It would be interesting to explore different ways to improve performance at few-shot classification. One way could be to compare the performance of meta-learning algorithms depending on the "shape" of the meta-training dataset. Would a dataset with 100 different classes and 500 examples per class allow better performance than a dataset with 50 classes and 1000 examples per class? My assumption is that it would, since it would allow the algorithm to better generalize to new classes, but this still needs to be proven.

Finally, in addition to the classification accuracy, it would be interesting to collect more intelligence about how meta-learning algorithm actually learn, for instance by studying the features representations, or using Explainable Machine Learning techniques, adapted to the meta-learning paradigm.

\subsection{MAML for Few-Shot object detection} \label{YOLOMAML}
\subsubsection{The Few-Shot Object Detection problem}
We saw that in real world applications, we sometimes need to solve an image classification task with only few examples. The same problem is encountered in all other tasks composing the field of computer vision. Here, we tackle the Few-Shot Object Detection problem.

\begin{figure}[h]
    \centering
    \includegraphics[width=10cm]{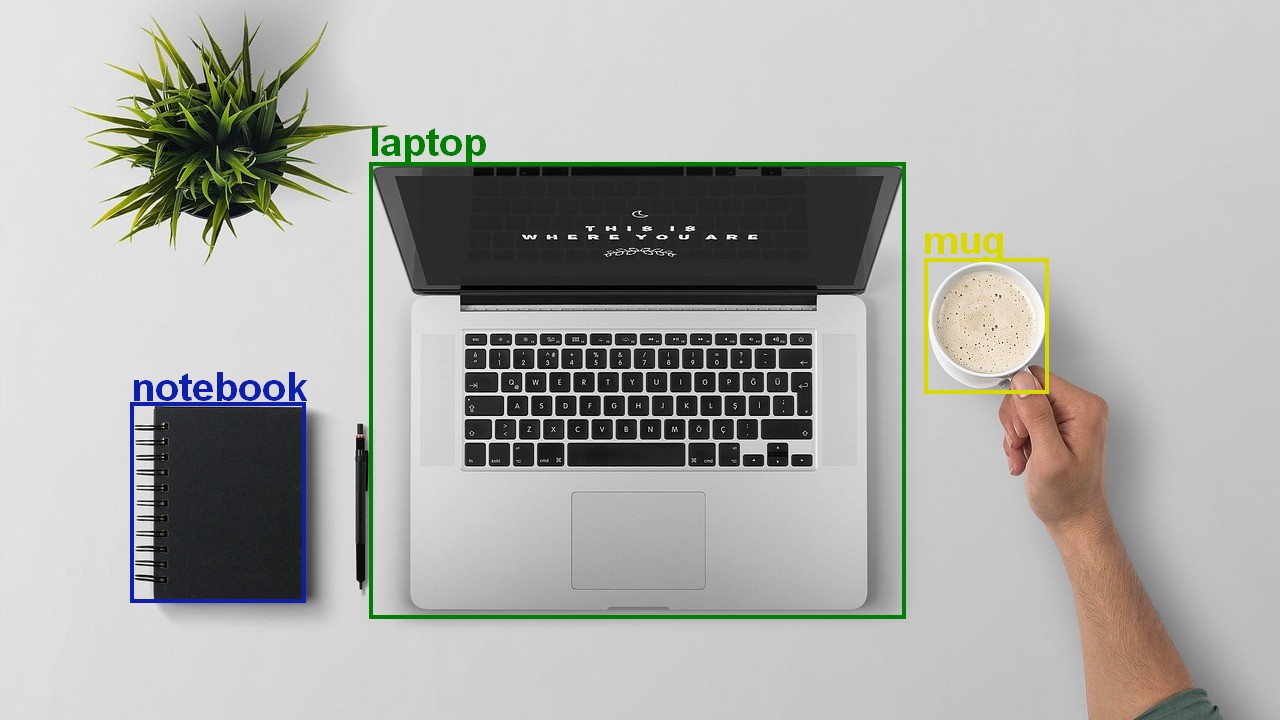}
    \caption{Consider the task of detecting objects belonging in $\lbrace$ laptop, mug, notebook, lamp $\rbrace$. Here the object detector detects and classifies a laptop, a mug and a notebook. It doesn't detect the plant and the pen since they are not part of the given task.}
    \label{fig:object-detection}
\end{figure}

Here we define the object detection task as follow: given a list of object types and an input image, the goal is to detect all object belonging in the list. Detecting an object consists in:
\begin{enumerate}
    \item localizing the object by drawing the smallest bounding box containing it;
    \item classifying the object.
\end{enumerate}
As such, object detection is the combination of a regression task and a classification task. An example is shown in Figure \ref{fig:object-detection}.

Following this, we define a $N$-way $K$-shot object detection task as follows. Given:
\begin{enumerate}
    \item a support set composed of:
    \begin{itemize}
        \item $N$ class labels;
        \item For each class, $K$ labeled images containing at least one object belonging to this class;
    \end{itemize}
    \item $Q$ query images;
\end{enumerate}
we want to detect in the query images the objects belonging to one of the $N$ given classes. The $N \times K$ images in the support set contain the only examples of object belonging to one of the $N$ classes.

When $K$ is small, we talk about \emph{few-shot object detection}.

We can immediately spot a key difference with few-shot image classification: one image can contain multiple objects belonging to one or several of the $N$ classes. Therefore, when solving a $N$-way $K$-shot detection tasks, the algorithm trains on \emph{at least} $K$ example objects for each class. During a $N$-way $K$-shot classification tasks, the algorithms sees \emph{exactly} $K$ examples for each class. Note that this can become a challenge: in this configuration, the support set may be unbalanced between classes. As such, this formalization of the few-shot object detection problem leaves room for improvement. It was chosen because it is a rather straightforward setup, which is also convenient to implement, as we will see in section \ref{yolomaml-implementation}.

\subsubsection{YOLOMAML}

To solve the few-shot object detection problem, we had the idea of applying the Model-Agnostic Meta-Learning algorithm~\cite{Finn17} to the YOLO~\cite{redmon2018yolov3} detector. We call it YOLOMAML for lack of a better name.

As presented in section \ref{gradient-based-meta-learning}, MAML can be applied to a wide variety of deep neural networks to solve many few-shot tasks. Finn \emph{et al.} considered few-shot classification and regression as well as reinforcement learning. It could as well be applied to a standard detector to solve few-shot object detection.

YOLOv3 is already used on other projects at Sicara. Our expertise on this detector motivated our choice to use it. Also, it prevents the advantage of being a single-stage detector. It appeared easier to apply MAML to YOLO than to a variant of R-CNN.

At the same time, Fu \emph{et al.} proposed the Meta-SSD~\cite{fu2019meta}. It applies the Meta-SGD~\cite{li2017meta} (a variant of MAML which additionally meta-learns hyper-parameters of the base model) to the Single-Shot Detector~\cite{liu2016ssd}. Fu \emph{et al.} presented promising results. Although Meta-SSD and YOLOMAML are very similar, I argue that it is relevant to continue to work on YOLOMAML, in order to:
\begin{enumerate}
    \item confirm or challenge the interesting results of Fu \emph{et al.}, with a similar algorithm and on a wider variety of datasets;
    \item disclose the challenges of developing such an algorithm.
\end{enumerate}

YOLOMAML is a straightforward application of the MAML algorithm to the YOLO detector. The algorithm is shown in Algorithm \ref{yolomaml-algorithm}.
\begin{algorithm}[h]
 \caption{YOLOMAML}
 \label{yolomaml-algorithm}
 \begin{algorithmic}[1]
    \REQUIRE $\alpha, \beta$, respectively the inner loop and outer loop learning rate
    \REQUIRE \texttt{n\_episodes} the number of few-shot detection tasks considered before each meta-gradient descent
    \REQUIRE \texttt{number\_of\_updates\_per\_task} the number of inner loop gradient descents in each few-shot detection task
    \STATE initialize the parameters $\theta$ of the YOLO detector $f_\theta$
    \WHILE{\textbf{not done}}
        \STATE sample \texttt{n\_episodes} detection tasks $\mathcal{T}_i$, where each task is defined by a support set $S_i=\lbrace x^S_j,l^S_j \rbrace$ and a query set $Q_i={x^Q_j,l^Q_j}$
        \FOR{$\mathcal{T}_i$ \textbf{in} $\lbrace\mathcal{T}_i\rbrace$}
            \STATE $\theta_0 \leftarrow \theta$
            \FOR{$t < \texttt{number\_of\_updates\_per\_task}$}
                \STATE compute the gradient of the loss of the YOLO model $f_{\theta_t}$ on the support set: $\nabla_{\theta_t} \mathcal{L}\left( f_{\theta_t}(\lbrace x^S_j \rbrace),\lbrace l^S_j \rbrace \right)$
                \STATE update $\theta_{t+1} \leftarrow \theta_t - \alpha \nabla_{\theta_t} \mathcal{L}\left( f_{\theta_t}(\lbrace x^S_j \rbrace),\lbrace l^S_j \rbrace \right)$
            \ENDFOR
            \STATE compute the gradient of the loss of the YOLO model $f_{\theta_\texttt{number\_of\_updates\_per\_task}}$ on the query set relative to initial parameters $\theta$: $\nabla_{\theta} \mathcal{L}\left( f_\texttt{number\_of\_updates\_per\_task}(\lbrace x^Q_j \rbrace),\lbrace l^Q_j \rbrace \right)$
        \ENDFOR
        \STATE update $\theta \leftarrow \theta - \beta \sum_{\mathcal{T}_i \in \lbrace\mathcal{T}_i\rbrace} \nabla_{\theta} \mathcal{L}\left( f_{\theta_\texttt{number\_of\_updates\_per\_task}}(\lbrace x^{Q_i}_j \rbrace),\lbrace l^{Q_i}_j \rbrace \right)$
    \ENDWHILE
 \end{algorithmic}
\end{algorithm}
 
\subsubsection{Implementation} \label{yolomaml-implementation}
I decided to re-use the structure of the MAML algorithm from my work on Image Classification. For the YOLO model, I used the implementation from Erik Linder-Norén \footnote{\url{https://github.com/eriklindernoren/PyTorch-YOLOv3}}, which is mostly a PyTorch reimplementation of Joseph Redmon's original C implementation \footnote{\url{https://github.com/pjreddie/darknet}}. It contains two main parts:
\begin{itemize}
    \item Data processing from raw images and labels to an iterable \texttt{Dataloader}. The class \texttt{ListDataset} is responsible for this process.
    \item The definition, training and induction of the YOLO algorithm, mostly handled by the class \texttt{Darknet}.
    \begin{itemize}
        \item It creates the YOLO algorithm as a sequence of PyTorch \texttt{Module} objects, from a configuration file customable by the user.
        \item It allows to load pre-trained parameters for part or all of the network.
        \item It defines the forward pass of the model and the loss computation.
    \end{itemize}
\end{itemize}

The experiences in few-shot object detection were made on the COCO 2014 dataset~\cite{lin2014microsoft}.

I had to work on three main levels of the implementation to allow complementarity between YOLO and MAML:
\begin{itemize}
    \item model initialization;
    \item fast adaptation of weights in convolutional layers
    \item data processing in the form of few-shot detection episodes
\end{itemize}

\paragraph{Model initialization} YOLOv3 in its standard form contains more than 8 millions parameters. Thus a full meta-training of it with MAML (which involves second order gradient computation) is prohibitive in terms of memory. Therefore:
\begin{enumerate}
    \item Instead of the standard YOLOv3 neural network, I used a custom Deep Tiny YOLO. The backbone of the model is the Tiny Darknet\footnote{\url{https://pjreddie.com/darknet/tiny-darknet/}}. On top of it, I added two output blocks (instead of three in the regular YOLOv3). The full configuration file of this network is available in the repository\footnote{\url{https://github.com/ebennequin/FewShotVision}} in \texttt{detection/configs/deep-tiny-yolo-5-way.cfg}.
    \item I initialized the backbone with parameters trained on ImageNet, then froze those layers.
\end{enumerate}
This way, there were only five trainable convolutional blocks left in the network. This allows to train the YOLOMAML on a standard GPU in a few hours. Note that there exists a Tiny YOLO, but there is no available backbone pre-trained on ImageNet for this network, which motivated my choice of a new custom network.

\paragraph{Fast adaptation} The core idea of MAML is to update the trainable parameters on each new task, while training the initialization parameters across tasks. For this, we need to store the updated parameters during a task, as well as the initialization parameters. A solution for this is to add to each parameter a field \texttt{fast} which stores the updated parameters. In our implementation (inherited from~\cite{Chen19}), this is handled by \texttt{Linear\_fw}, \texttt{Conv2d\_fw} and \texttt{BatchNorm2d\_fw}  which respectively extend the \texttt{nn.Linear}, \texttt{nn.Conv2d} and \texttt{nn.BatchNorm2d} PyTorch objects. I modified the construction of the \texttt{Darknet} objects so that they use these custom layers instead of the regular layers.

\paragraph{Data processing} As in few-shot image classification, we can sample a $N$-way $K$-shot detection task with $Q$ queries per class by first sampling $N$ classes. Then, for each class, we sample $K+Q$ images which contain at least one box corresponding to this class. The difference in detection is that we then need to eliminate from the labels the boxes that belong to a class that does not belong to the detection task. There would be the same problem with multi-label classification.

To solve this problem, I created an extension of the standard PyTorch \texttt{Sampler} object: \texttt{DetectionTaskSampler}. In addition to returning the indices of the data instances to the DataLoader, it returns the indices of the sampled classes. This information is processed in \texttt{ListDataset} to feed the model proper few-shot detection task with no reference to classes outside the task.

\subsubsection{First results and investigations}
My attempts to build a working few-shot object detector are to this day unsuccessful. In this section, I will expose my observations and attempts to find the source(s) of the problem. 

I launched a first experiment with a Deep Tiny YOLO initialized as explained in the previous section. It is trained on 3-way 5-shot object detection tasks on the COCO dataset. It uses an Adam optimizer with a learning rate of $10^{-3}$ (both in the inner loop and outer loop). It is trained for 10 000 epochs, each epoch corresponding to one gradient descent on the average loss on 4 episodes. During each episode, the model is allowed two updates on the support set before performing detection on the query set.

\begin{figure}[h]
    \centering
    \includegraphics{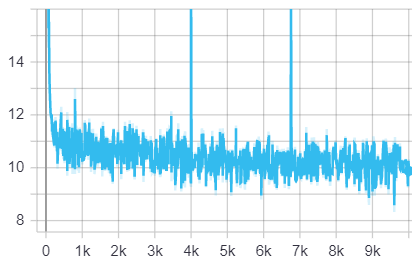}
    \caption{Total loss per epoch of YOLOMAML. Each point is the average total loss of the model on the query set of the episodes of one epoch.}
    \label{fig:tensorboard1}
\end{figure}

The loss is quickly converging (see Figure \ref{fig:tensorboard1}) but at inference time, the model is unable to perform successful detections (with a F1-score staying below $10^{-3}$). Extensive hyperparameter tuning has been performed with no sensible improvement on the results.

\begin{figure}[h]
  \begin{minipage}[b]{0.5\linewidth}
   \centering
   \includegraphics[width=8cm]{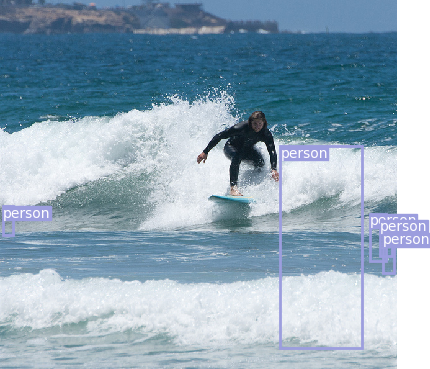}      
  \end{minipage}
\hfill
  \begin{minipage}[b]{0.48\linewidth}
   \centering
   \includegraphics[width=8cm]{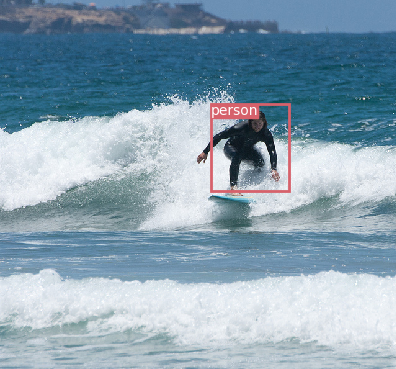}      
  \end{minipage}
  \begin{minipage}[b]{0.5\linewidth}
   \centering
   \includegraphics[width=8cm]{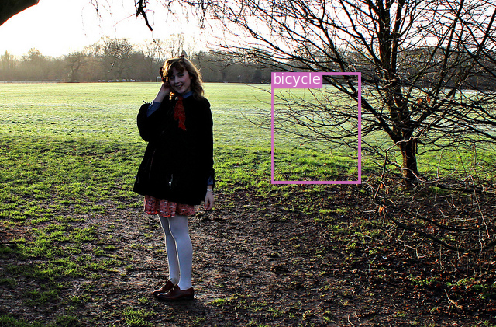}      
  \end{minipage}
\hfill
  \begin{minipage}[b]{0.48\linewidth}
   \centering
   \includegraphics[width=8cm]{y_COCO_val2014_000000161875.png}      
  \end{minipage}

  \caption{Object detection by the models YOLOMAML (left column) and YOLO (right column).}
  \label{fig:inference}

\end{figure}

To ensure that these disappointing performance was not due to my reimplementation of YOLO, I trained the Deep Tiny YOLO without MAML, in the same settings, for 40 epochs. Although this training is not optimal, the model is still able to perform relevant detections, which is not the case for YOLOMAML (see Figure \ref{fig:inference}).

The YOLOv3 algorithm aggregates three losses on three different parts of the predictions:
\begin{enumerate}
    \item the shape and position of the bounding box of predicted objects, using Mean Square Error;
    \item the objectness confidence (how sure is the model that there is truely an object in the predicted bounding box) using Binary Cross Entropy;
    \item the classification accuracy on each predicted box, using Cross Entropy.
\end{enumerate}

\begin{figure}[h]
    \centering
    \includegraphics[width=15cm]{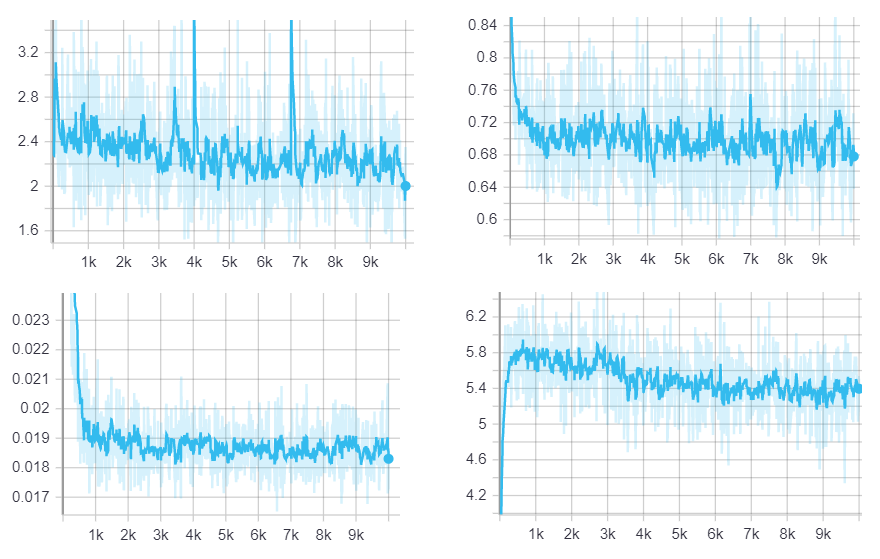}
    \caption{Evolution of the four parts of the loss of YOLOMAML during the same training as in Figure \ref{fig:tensorboard1}. Up-left: bounding box loss. Up-right: classification loss. Bottom-left: objectness confidence loss for boxes with no ground truth object. Bottom-right: objectness confidence loss for boxes with a ground truth object. Exponential moving average has been used to clearly show the patterns.}
    \label{fig:tensorboard2}
\end{figure}

Figure \ref{fig:tensorboard2} shows the evolution of these different parts of the loss. Loss due to objectness confidence has been further divided into two parts : the loss on boxes that contain an object in the ground truth, and the loss on boxes that do not contain an object in the ground truth.

We can see that the loss due to the classification and to the shape and position of the bounding box do not evolve during training. The no-object-confidence loss drops in the first thousand epochs before stagnating, while the yes-object-confidence rises to a critical amount before stagnating.

\begin{figure}[h]
    \centering
    \includegraphics[width=15cm]{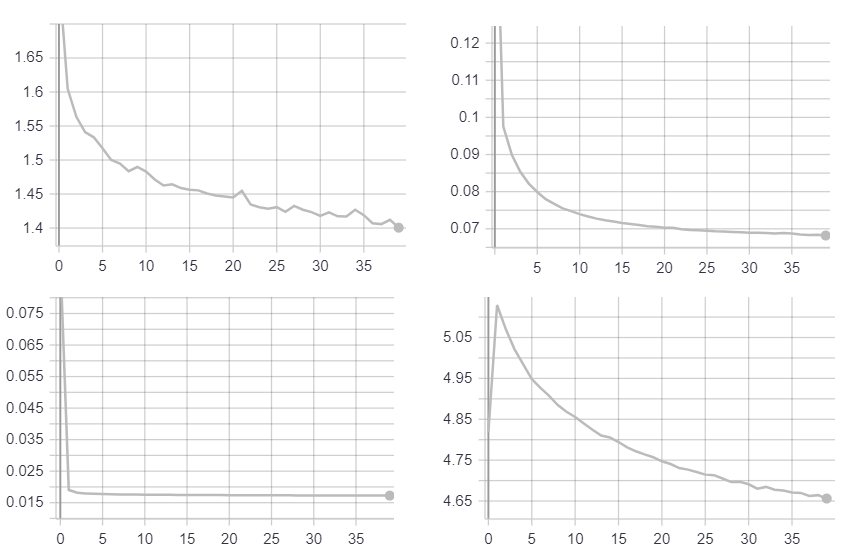}
    \caption{Evolution of the four parts of the loss of YOLO. Up-left: bounding box loss. Up-right: classification loss. Bottom-left: objectness confidence loss for boxes with no ground truth object. Bottom-right: objectness confidence loss for boxes with a ground truth object.}
    \label{fig:tensorboard3}
\end{figure}

Figure \ref{fig:tensorboard3} shows the same data for the training of YOLO. We can see that in this case, the yes-object-confidence drops after a peak in the first epochs. All parts of the loss decrease during the training, except the no-object-confidence, which reaches a floor value which is relatively small compared to the other parts.

Considering this, it is fair to assume that the bottleneck in training YOLOMAML is the prediction of the objectness confidence.

\subsubsection{Future work}
Unfortunately I did not have enough time to develop a working version of YOLOMAML. At this point I believe the answer resides in the prediction of the objectness confidence, but it is likely that other issues may rise when this one is solved. 

An other direction of future work would be to constitute a dataset adapted to few-shot detection. Other works~\cite{kang2018few}~\cite{fu2019meta} propose a split of the PASCAL VOC dataset adapted to few-shot detection. However, PASCAL VOC contains only 25 classes, while COCO contains 80 classes. I believe this makes COCO more adapted to meta-learning, which is entangled with the idea of learning to generalize to new classes.

Finally, a drawback of a (working) YOLOMAML would be that it does not allow way change, \emph{i.e.} that a model trained on $N$-way few-shot detection tasks cannot be applied to a $N'$-way few-shot detection tasks. Solving this problem would be a useful improvement for YOLOMAML.

\newpage
\section{Conclusion}
Advanced research in Few-Shot Learning is still young. Until now, only few works have tackled the few-shot object detection problem, for which there is yet no agreed upon benchmark (like mini-ImageNet for few-shot classification). However, solving this problem would be a very important step in the field of computer vision. Using meta-learning algorithms, we could have the ability to learn to detect new, unseen objects with only a few examples and a few minutes.

I am disappointed that I was not able to make YOLOMAML work during my internship at Sicara. However, I strongly believe that it is important to keep looking for new ways of solving few-shot object detection, and I intend to keep working on this.

\newpage

\bibliographystyle{unsrt}
\bibliography{mybib}

\end{document}